%% file: paper.tex
\documentclass[11pt]{article}
\usepackage{coling2018}
\usepackage{times}
\usepackage{url}
\usepackage{latexsym}
\usepackage{graphicx}
\usepackage{wrapfig}
\usepackage{amsmath}
\usepackage{amssymb}
\usepackage{mathtools}
\usepackage{xcolor}
\usepackage{placeins}

\newcommand{\svdpmi}{SVD$_\text{PPMI}$}
\newcommand{\jeseme}{\textsc{JeSemE}}

\title{\jeseme: A Website for Exploring Diachronic Changes \\in Word Meaning and Emotion}

\author{Johannes Hellrich$^{1,2}$ \hspace{25pt}  Sven Buechel$^2$ \hspace{25pt} Udo Hahn$^2$\vspace{8pt}\\
$^1$ Graduate School `The Romantic Model'\\ 
\url{modellromantik.uni-jena.de}
\vspace{4pt}\\
$^2$ Jena University Language \& Information Engineering (JULIE) Lab\\
\url{julielab.de}
\vspace{4pt}\\
Friedrich-Schiller-Universit{\"a}t Jena, Jena, Germany
}

\begin{document}
\maketitle
\begin{abstract}
We here introduce a substantially extended version of \jeseme, an interactive website for visually exploring computationally derived time-variant information on word meanings and lexical emotions assembled from five large diachronic text corpora. \jeseme\ is designed for scholars in the (digital) humanities as an alternative to consulting manually compiled, printed dictionaries for such information (if available at all). This tool uniquely combines state-of-the-art distributional semantics with a nuanced model of human emotions, two information streams we deem beneficial for a data-driven interpretation of texts in the humanities.
\end{abstract}

%
%
\blfootnote{
     \hspace{-0.65cm}  
     This work is licensed under a Creative Commons 
     Attribution 4.0 International License.
     License details:
     \url{http://creativecommons.org/licenses/by/4.0/}
}


\input{intro}

\input{architecture}

\input{emotions}

\input{example}
\input{conclusion}

\section*{Acknowledgements}
This research was partially funded by grant GRK 2041/1 from \textit{Deutsche For\-schungs\-gemeinschaft} within the Graduate School
 \textit{``The Romantic Model.\ Variation--Scope--Relevance''}.

\bibliographystyle{acl}
\bibliography{literature-short}
\end{document}

%% file: intro.tex
\section{Introduction}
\label{intro}

Historical, manually compiled dictionaries are central to many kinds of studies in the humanities, since they provide scholars with information about the lexical meaning of terms in former time periods. Yet, this traditional approach is limited in many ways, coverage being perhaps the most pressing issue: Is a dictionary for the specific time period a scholar is investigating really available and, if so, does it cover all of the lexical items of interest?

Word embeddings have been proposed as a technical vehicle to increase lexical coverage \cite{Kim14}. However, they require locally installed software and time-consuming calculations, thus being ill-suited for mostly non-technical users in the humanities. As an alternative, we here present an extended version of \jeseme , a user-friendly open source website\footnote{
	Website available at:\url{jeseme.org}; sources available at: \url{github.com/JULIELab/JeSemE}} 
for accessing embedding-derived diachronic information on lexical meaning and emotion.
The first release of \jeseme\ \cite{Hellrich17acl} mainly provided  time-variant diachronic lexical \mbox{semantic} information. Its second version, the focus of this paper, excels with the unique capability to additionally track the diachronic \textit{\mbox{emotional}} connotation of words in parallel with their lexical semantics. Such a functionality is widely considered beneficial for the data-driven interpretation of literary text genres \cite{Kim17}.

Measuring affective information on the lexical level is an active field of research in computational linguistics \cite{Liu15}. Yet, most contributions focus on contemporary language and are limited to shallow representations of human emotions, mainly distinguishing between \textit{positive}  and \textit{negative} feelings. 
Current research in sentiment analysis either starts to include historical trends in word polarity \cite{Hamilton16emnlpEmo} or incorporates more nuanced models of emotions, such as Valence-Arousal-Dominance \cite{Buechel18naacl}. This contribution integrates both lines of work in a unique way based on our prior research activities \cite{Buechel16lt4dh,Buechel17dh}.
To the best of our knowledge, only few systems share similarities with \jeseme. Alternative websites for tracking diachronic word meaning yet offer far less diverse collections of corpora compared to \jeseme\ and neither of them incorporates emotion values attached to lexical entries. 
For example, \newcite{Arendt17} provide only short term trends in word similarity in two social media corpora in their \textsc{Esteem} system.\footnote{\url{esteem.labworks.org/}} The system\footnote{\url{embvis.flovis.net/s/neighborhoods.html}} by \newcite{Heimerl18} is intended as a mere showcase for a novel visualization technique and re-uses SGNS embeddings trained on the English Google Books corpus by \newcite{Hamilton16acl}. The \textsc{Diachronic Explorer}\footnote{\texttt{tec.citius.usc.es/explorador-diacronico}} which uses sparse vector representations instead of word embeddings to calculate lexical similarity is limited to the Spanish Google Books corpus \cite{Gamallo17cee}.

%% file: architecture.tex
\section{Architecture and Website}

\begin{figure}[t]\vspace*{-5pt}
\center
\fbox{
\includegraphics[width=.85\textwidth]{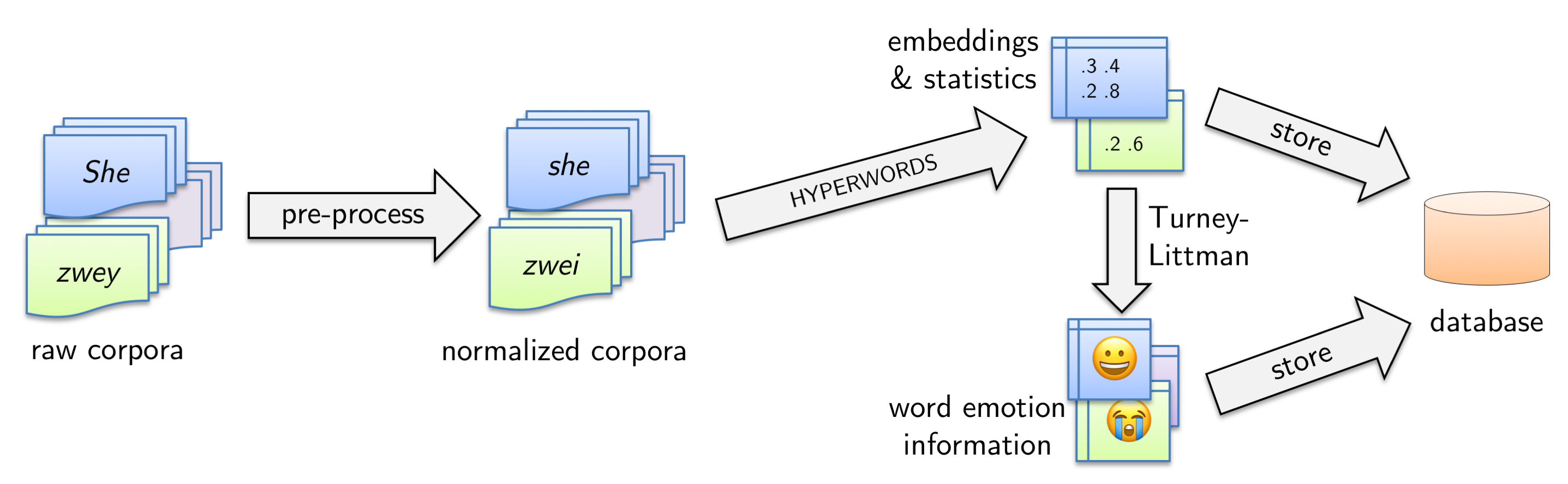} \vspace*{-25pt}
}
\caption{\jeseme 's text processing pipeline.}\label{fig:pipeline} \vspace*{-5pt}
\end{figure}

\begin{wrapfigure}{r}{0.63\textwidth}
\vspace*{-5pt}
\centering
\fbox{
\includegraphics[width=0.6\textwidth]{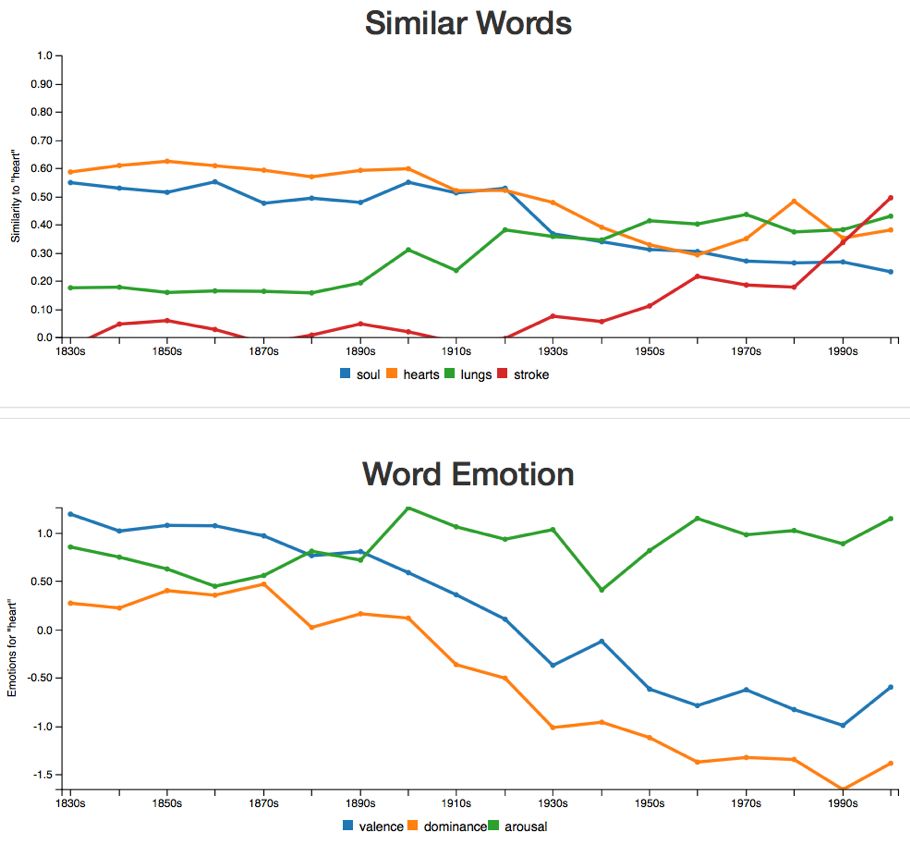} 
}\vspace*{-5pt}
\caption{\jeseme\ in operation I: Meaning change of ``heart'' relative to reference words since the 1830s in the COHA. 
}
\end{wrapfigure}\label{fig:screenshots}

\jeseme\ uses five diachronic corpora: the Google Books N-Gram Corpus for German and its English fiction register  \cite{Michel11}, the Corpus of Historical American English (COHA; \newcite{Davies12}), the Deutsches Textarchiv [`German Text Archive'] \cite{Geyken13} and the Royal Society Corpus \cite{Kermes16}. To ensure high embedding quality, these corpora are divided into temporal slices of  similar size covering between 10 to 50 years each.

\jeseme 's processing pipeline is depicted in Figure \ref{fig:pipeline}. It starts with orthographically normalizing the corpus slices, i.e., lower casing only for  English and a historical spelling-aware lemmatization for  German \cite{Jurish13}. We then use a modified version of \textsc{Hyperwords}\footnote{
	\url{github.com/hellrich/hyperwords}} 
to calculate slice-specific embedding models with \svdpmi \ \cite{Levy15}. This algorithm was chosen for its superior reliability which is essential for interpreting local neighborhoods in embedding spaces as is done in the remainder of this paper \cite{Hellrich16coling,Hellrich17dh}.
Apart from word vectors, we also calculate word-based co-occurrence statistics, frequency information and emotion values for each slice (see Section \ref{sec:emotions}). All this information is stored in a relational database.
Compared to \newcite{Hellrich17acl}, our current version also reduces the database size from approximately 120GB to 40GB. This is achieved by storing word vectors instead of pre-computed similarity scores. 
Unlike the previous version, semantic similarity between most words will  be computed on the fly. Only the most similar ones for each word (automatically picked as references) are cached for fast retrieval.

\jeseme 's website prompts a search form for selecting the word under scrutiny as well as one of the five corpora we supply. Its result page then provides graphs depicting the development of semantic similarity to automatically chosen reference words over time as an indicator for semantic change, as well as information on diachronic affective meaning (see Figure \ref{fig:screenshots}).
These two main sources of information are complemented with information on word co-occurrence and relative frequency, thus providing scholars with additional information to increase interpretability and rule out measurement artifacts.
Users may also add further reference words to the analysis on demand.
Besides this graphical interface \jeseme\ also offers a \texttt{REST} API.\footnote{See online documentation: \url{jeseme.org/help.html#api}}

%% file: emotions.tex
\section{Representing and Computing Emotions} 
\label{sec:emotions}

\begin{wrapfigure}{r}{0.63\textwidth}
\vspace*{-7pt}
\center
\fbox{
\includegraphics[width=0.6\textwidth]{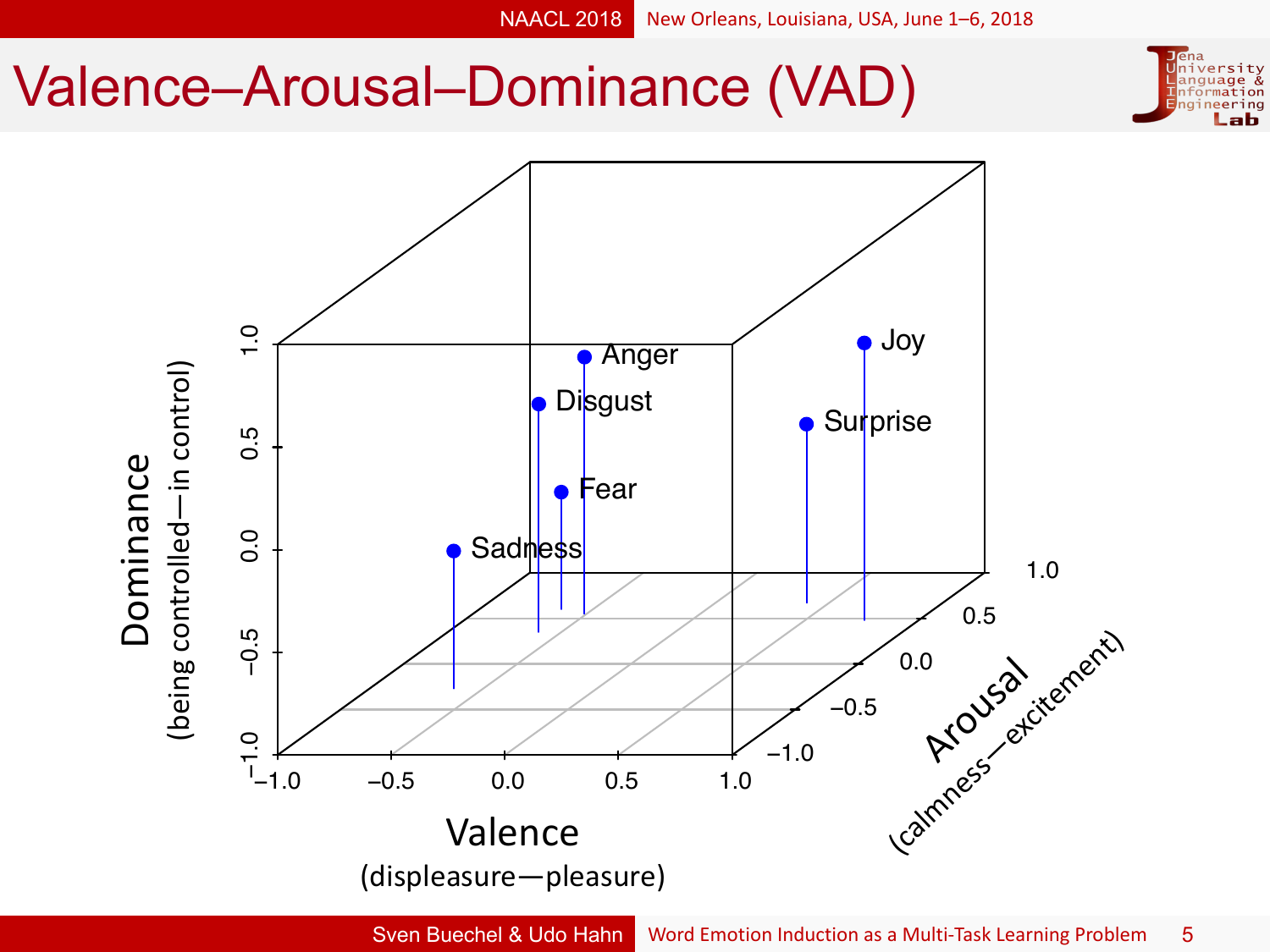} 
}\vspace*{-5pt}
\caption{Affective space spanned by the Valence-Arousal-Dominance (VAD) model, together with the position of six basic emotion categories. Adapted from \newcite{Buechel16ecai}. \label{fig:VAD}}
\vspace*{-5pt}
\end{wrapfigure}

We represent emotions following the Valence-Arousal-Dominance (VAD) scheme \cite{Bradley94}, one of the major models of emotion in psychology (for an illustration, see Figure \ref{fig:VAD}).
The VAD model describes affective states relative to three dimensions, namely, Valence (degree of displeasure vs.\ pleasure), Arousal (degree of calmness vs.\ excitement) and Dominance (degree of perceived control in a social situation).

We used a modified version of the emotion induction algorithm by \newcite{Turney03} based on evidence that it outperforms alternative methods for historical emotion lexicon creation \cite{Buechel17dh,Hellrich18arxiv}.
In this work, each word's predicted emotion value $\hat{e}(w)$ is calculated by averaging the emotion values $e(s)$ for each member $s$ of a seed set $S$, with $sim(w,s)$, the similarity between $w$ and $s$, serving as a weight:
\begin{equation*}
\hat{e}(w) := \frac{\sum_{s \in S} sim(w,s) \times e(s)}{\sum_{s \in S} sim(w,s)}
\end{equation*}

For the emotion scores stored in \jeseme, we used the emotion lexicons by \newcite{Warriner13} and \newcite{Schmidtke14} as seed sets for English and German corpora, respectively. 
Word emotions were induced independently for each temporal corpus slice, using the respective embedding model to retrieve similarity scores. 
Hence, the similarity between the seed words and the target word reflects word usage at a given language stage, thereby infusing historical emotion information into the resulting emotion ratings \cite{Buechel17dh}.

%% file: example.tex
\section{Examples}
\label{sec:results}

The new insights provided by diachronic emotion models can be demonstrated by re-visiting the example of ``heart'' we used in \newcite{Hellrich17acl} as shown in Figure \ref{fig:screenshots}. This lexeme  is often used metaphorically or metonymically despite the fact that the heart's anatomical function was already known for a long time. Results for our novel emotion tracking functionality match a move from metaphorical to anatomical usage we previously observed in the genre-balanced COHA. Around 1900, the similarity of ``heart'' to words such as ``stroke'' increases, while Dominance and Valence ratings drop sharply in tandem (see Figure \ref{fig:screenshots}; y-axis values are centered and scaled).
This simultaneous drop seems plausible, since we can ``change our heart'' in a metaphorical sense, yet have little control over our anatomical heart. Also, with its increasing anatomical usage, ``heart'' becomes less positive, since we are under mortal threat by cardiovascular diseases such as a ``stroke''.

\begin{wrapfigure}{r}{0.63\textwidth}
\centering
\vspace*{-5pt}
\fbox{
\includegraphics[width=0.6\textwidth]{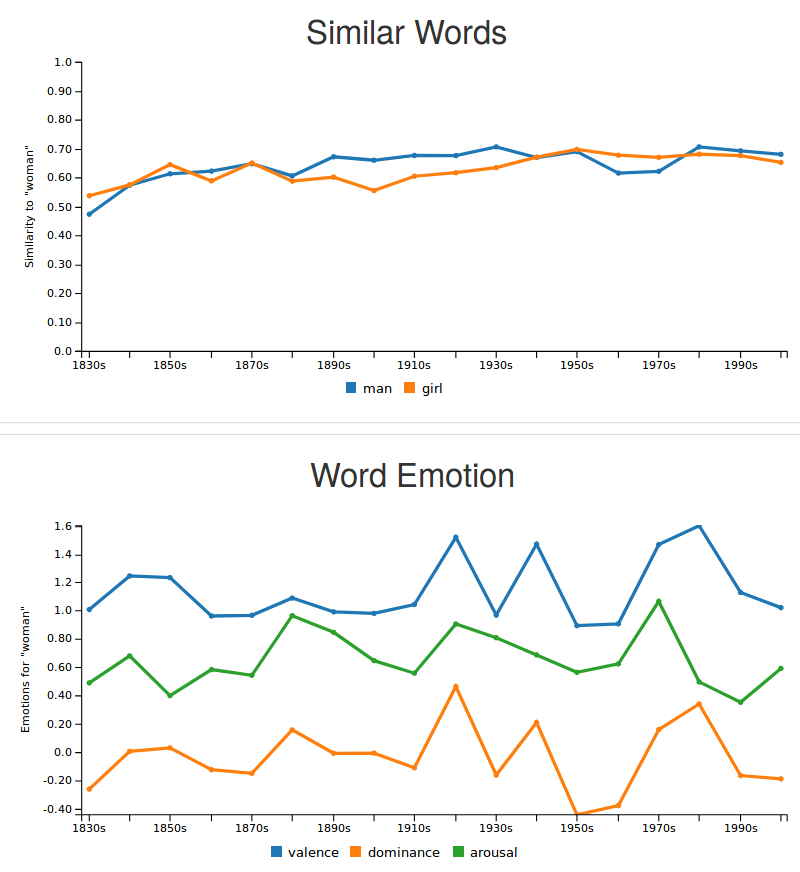} 
}\vspace*{-5pt}
\caption{\jeseme\ in operation II: Meaning of ``woman'' since the 1830s in the COHA.
	\vspace*{-25pt}
}
\end{wrapfigure}\label{fig:screenshot2}

Changes in emotion can also be traced for items with a more constant meaning, e.g., for ``woman'' as shown in Figure \ref{fig:screenshot2}. Here similarity scores for the most similar words---``man'' and ``girl''---remain rather static. Yet, emotion values are highly dynamic and seem to match turning points in women's rights movement, e.g., women's suffrage in the US is connected with an increase in all VAD dimensions for the 1920s.

\FloatBarrier

%% file: conclusion.tex
\section{Conclusion}

We introduced a substantially extended version of \jeseme, an interactive website for tracking diachronic changes in word meaning and, as a novel and unique feature, word emotion. 
To the best of our knowledge, no other system combines these two traits. 
\jeseme\ allows users with a limited technical background to interactively explore semantic evolution based on five large diachronic corpora for two languages, German and English. 
We believe that  \jeseme\ will be most useful for diachronic linguists and scholars within the digital humanities. We see two major applications: First,  it  can be used to generate hypotheses by querying words of interest to get a first impression of their semantic evolution. Second, scholars can first shape a hypothesis using traditional means and then query \jeseme\ for testing its plausibility based on diachronic statistical evidence.